\documentclass[10pt, a4paper]{article}
\usepackage{lrec2014/lrec2014}

\usepackage[justification=centering]{caption}
\usepackage{graphicx}
\usepackage{multirow}
\usepackage{acronym}
\usepackage{stfloats}
\usepackage{algorithm}
\usepackage{algpseudocode}
\usepackage{amsmath,amssymb,amsthm}
\usepackage{stmaryrd}
\usepackage{hhline}
\usepackage[table]{xcolor}

\usepackage[english]{babel}

\usepackage{pgf}
\usepackage{tikz}
\usetikzlibrary{automata,positioning,shapes,arrows}

\theoremstyle{theorem}

\newtheorem{definition}{Definition}
\newtheorem{example}{Example}

\newcommand{\ssl}{{\mathsf{SL}}}
\newcommand{\pdlsl}{$\mathrm{PDL}_\ssl$}
\newcommand{\pdlslblack}{$\mathrm{\mathbf{PDL}}_\mathsf{\mathbf{SL}}$}

\def\thesubsection{\arabic{section}.\arabic{subsection}}

\title{Implementation of an Automatic Sign Language Lexical Annotation Framework based on Propositional Dynamic Logic}

\name{Arturo Curiel$^{\dagger}$ \thanks{$^\dagger$ Supported by CONACYT (Mexico) scholarship program.}, Christophe Collet}

\address{ Universit\'e Paul Sabatier - Toulouse III\\
               118 route de Narbonne, IRIT,\\
               31062, Toulouse, France \\
               E-mail: curiel@irit.fr, collet@irit.fr\\}

\acrodef{sl}[SL]{sign language}
\acrodef{lsf}[FSL]{french sign language}
\acrodef{lts}[LTS]{labeled transition system}
\acrodef{mrsw}[MRSW]{multiple-reader/single-writer}
\acrodef{nlp}[NLP]{natural language processing}
\acrodef{pdl}[PDL]{Propositional Dynamic Logic}
\acrodef{bnf}[BNF]{Backus--Naur Form}
\acrodef{pdlsl}[\pdlsl]{Propositional Dynamic Logic for Sign Language}

\newcommand{\signlsf}[1]{$\mathtt{#1}_\mathsf{FSL}$}

\abstract{In this paper, we present the implementation of an automatic \ac{sl} sign annotation framework based on a formal logic, the \ac{pdl}. Our system relies heavily on the use of a specific variant of \ac{pdl}, the \ac{pdlsl}, which lets us describe \ac{sl} signs as formulae and corpora videos as \acp{lts}. Here, we intend to show how a generic annotation system can be constructed upon these underlying theoretical principles, regardless of the tracking technologies available or the input format of corpora. With this in mind, we generated a development framework that adapts the system to specific use cases. Furthermore, we present some results obtained by our application when adapted to one distinct case, 2D corpora analysis with pre-processed tracking information. We also present some insights on how such a technology can be used to analyze 3D real-time data, captured with a depth device.
\newline\newline \Keywords{sign language framework, automatic annotation, propositional dynamic logic}}

\begin{document}

\maketitleabstract
\acresetall
\section{Introduction}

    Research in \ac{sl}, both from the point of view of linguistics and computer science, relies heavily on video-corpora analysis~\cite{dreuw_spoken_2008}. As such, several methods have been developed over time for the automatic processing of both video or other sensor-based corpora~\cite{ong_automatic_2005}. Even though these kind of research efforts are usually geared toward recognition, few work has been done in relation to the unification of raw tracked data with high level descriptions~\cite{moeslund_slrecognition_2011,bossard_issues_2004}. This calls to a reflection on how we represent \ac{sl} {\em computationally}, from the most basic level.

    \ac{sl} lexical representation research is focused on sign synthesis before than recognition. Works like~\cite{filhol_zebedee:_2009,losson_sign_1998} present the use of geometric lexical descriptions to achieve animation of signing 3D avatars. While their approach is well suited for synthesis, it is not completely adapted for sign identification. Recognition tasks in both natural language processing and computer vision are well known to be error-prone. Also, they are highly susceptible of bumping into incomplete information scenarios which may require some kind of inference, in order to effectively resolve ambiguities. In addition, \ac{sl} linguistic research has consistently shown the existence of common patterns across different \acp{sl}~\cite{aronoff_paradox_2005,meir_re-thinking_2006,wittmann_classification_1991} that may be lost with the use of purely geometrical characterizations, as the ones needed in synthesis. This limits the application of these kind of sign representations for automatic recognition, especially since we would want to exploit known linguistic patterns by adding them as properties of our descriptions. Works like~\cite{kervajan_french_2006,dalle_high_2006} have acknowledged the necessity of introducing linguistic information to enrich interaction, in an effort to help automatic systems bear with ambiguity. Moreover, the use of additional linguistic data could simplify connections between lexical information and higher syntactic-semantic levels, hence pushing us closer to automatic discourse analysis. However, this has long been out of the scope of synthesis-oriented description languages. 
    
    On the side, research in \ac{sl} recognition has to deal with other important drawbacks not present in synthesis, namely the use of very specialized tools or very specific corpora. This alone can severely impact the portability of a formal, computer-ready, representation out of the original research context, as it complicates the use of the same techniques across different information sources and toughens integration with new tools. 

    The framework described here is based on previous work presented by \cite{curiel_logic_2013} on the \ac{pdlsl}. \ac{pdlsl} is a formal logic created with the main purpose of representing \ac{sl} signs in a computer-friendly way, regardless of the specific tools or corpora used in research. Such a representation can potentially reduce the overhead of manually describing \ac{sl} signs to a computer, by establishing well-known sets of rules that can be interpreted by both humans and automatic systems. This could, incidentally, reduce dependency on thoroughly geometrical descriptions. Moreover, the flexibility of \ac{pdlsl} lets us combine any kind of information in our descriptions; for example, we can integrate non-manual markers if we have sight and eyebrow tracking, or we can add 3D movements if we are using a depth camera.
    
    In general, we propose an automatic \ac{sl} lexical annotation framework based in \ac{pdlsl} descriptions. Ideally, the system will:
    
    \begin{itemize}
    \item simplify the application of logical inference to recognize \ac{pdlsl}-described signs;
    \item characterize and analyze corpora in terms of \ac{pdlsl} models;
    \item represent \ac{sl} with different degrees of granularity, so as to adapt the formulae to the specific technical capabilities available in each use case.
    \end{itemize}
    
    Our framework aims to ease the integration of \ac{pdlsl} with various corpora and tracking technologies, so as to improve communication between different \ac{sl} research teams. We expect that this will, in turn, enable the construction of both research and user-level applications in later stages.
    
    The rest of the paper is divided as follows. In section~\ref{sec:formalmodel}, we introduce the basic notions of our formal language, applied to 2D \ac{sl} video-corpora analysis. Section~\ref{sec:pdlslformulae} shows how we can describe \ac{sl} lexical structures as verifiable \ac{pdlsl} formulae. Section~\ref{sec:systemarch} gives a detailed description of the system's architecture. Finally, sections \ref{sec:results} and \ref{sec:conclusions} present some preliminary results and conclusions, respectively.

\section{Sign Language Formalization with Logic}
    \label{sec:formalmodel}

    The \ac{pdl} is a multi-modal logic first defined by \cite{fischer_propositional_1979} to characterize computer languages. Originally, it provided a formal framework for program descriptions, allowing them to be interpreted as modal operators. \ac{pdlsl} is an specific instance of \ac{pdl}, based on the ideas of sign decomposition by \cite{liddell_american_1989} and \cite{filhol_modedescriptif_2008}. In general, \ac{pdlsl}'s modal operators are movements executed by {\em articulators}, while static postures are interpreted as propositional states reachable by chains of movements. 
        
    A propositional state will be none other than a set of distinct atomic propositions. These can be used to represent articulators' positions with respect to one another; specific configurations; or even their spatial placement within a set of {\em places of articulation}. Table~\ref{model:propositions} shows a brief summary of the atomic propositions defined to analyze 2D corpus data.
    
    \begin{table}[htb]
        \centering{
	    \begin{tabular}{|c|p{.7\columnwidth}|}
		    \hline
		    \textbf{Symbol}& \textbf{Meaning}\\ \hline
            ${\beta_1}^{\delta}_{\beta_2}$ & articulator ${\beta_1}$ is placed in relative direction $\delta$ with respect to articulator ${\beta_2}$.\vspace{.8mm}\\ 
            $\mathcal{F}^{\beta_1}_c$ & articulator ${\beta_1}$ holds configuration $c$.\vspace{1.2mm}\\
            $\Xi^{\beta_1}_\lambda$ & articulator ${\beta_1}$ is located in articulation place $\lambda$.\\
            $\mathcal{T}^{\beta_1}_{\beta_2}$ & articulator ${\beta_1}$ and ${\beta_2}$ touch.\\
		    \hline
	    \end{tabular}
	    \caption{Atomic propositions for \ac{pdlsl}}
	    \label{model:propositions}}
    \end{table}
    
    Basic movements can be described by atomic actions codifying either their direction, speed or even if they follow a particular trajectory. This is exemplified by the definitions on Table~\ref{model:actions}, which presents some of the operators used to characterize 2D corpus movements.
    
        \begin{table}[htb]
            \centering{
    	    \begin{tabular}{|c|p{.7\columnwidth}|}
    		    \hline
    		    \textbf{Symbol}& \textbf{Meaning}\\ \hline
                $\delta_{\beta_1}$ & articulator ${\beta_1}$ moves in relative direction $\delta$. \\
                $\leftrightsquigarrow_{\beta_1}$ & articulator ${\beta_1}$ {\em trills}, moves rapidly without direction. \\
                $\mathbf{skip}$ & denotes the execution of any action \\
    		    \hline
    	    \end{tabular}
    	    \caption{Atomic actions for \ac{pdlsl}}
    	    \label{model:actions}}
        \end{table}

    Both atomic propositions and actions presented in this case were chosen specifically to capture information that we are able to detect with our tracking tools. Different sets of atoms can be defined depending of the technical capabilities available to assert their truth values ({\em e.g.} sight direction, eyebrow configuration, hand movement, etc).

    Atoms form the core of the \ac{pdlsl} language, which is presented below in \ac{bnf} by way of definitions~\ref{def:actionlanguage}~and~\ref{def:pdlsl}.
    
   \begin{definition}[Action Language for SL Body Articulators $\mathcal{A}_\mathsf{SL}$]\label{def:actionlanguage}
                 
       \[\alpha::=\pi~|~\alpha\cap\alpha~|~\alpha\cup\alpha~|~\alpha;\alpha~|~\alpha^*\]
            where $\pi$ is an atomic action.
   \end{definition}
   
   \begin{definition}[Language \pdlsl]\label{def:pdlsl}

       \[\varphi ::= \top~|~p~|~\neg\varphi~|~\varphi \wedge \varphi~|~[\alpha]\varphi\]
       
       where $p$ denotes an atomic proposition and $\alpha \in \mathcal{A}_\mathsf{SL}$.
   \end{definition}
   
   A more formal presentation of the model basis can be found in \cite{curiel_logic_2013}.

\section{Extending \pdlslblack ~formulae to Describe Sign Language Lexical Properties}
    \label{sec:pdlslformulae}

     The presented \ac{pdlsl} language lets us easily codify individual signs by way of our logic formulae. However, during implementation, we noticed the need to extend the original formalism in order to develop a better suited characterization of more general properties. We wanted to represent lexical structures common across multiple signs. With this in mind, we extended \ac{pdlsl} to include {\em lambda expressions}, explained in~\cite{barendsen_introduction_1994}, for variable binding. The introduced syntax is presented in definition~\ref{def:lambdapdlsl}.

     \begin{definition}[Extended \pdlsl]\label{def:lambdapdlsl}

       \[var ::= \mathtt{\langle uniqueID \rangle}~|~var,var \]
       \[\varphi_f ::= \varphi~|~var~|~\neg \varphi_f~|~\varphi_f\wedge\varphi_f~|~\lambda~var.(\varphi_f)~|~var=\varphi_f\]
       
       where $\varphi \in~$\pdlsl.
     \end{definition}

    The rules of quantification and substitution remain the same as in classic lambda calculus.
    
\begin{figure}[htb]
        \centerline{
            \begin{minipage}{0.4\columnwidth}
                \begin{center}
                    \includegraphics[width=\columnwidth]{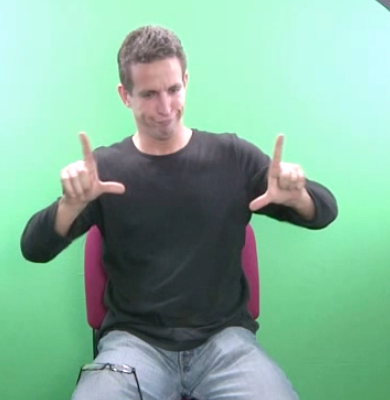}
                    \signlsf{SCREEN}
                \end{center}
            \end{minipage}
            \begin{minipage}{0.4\columnwidth}
                \begin{center}
                    \includegraphics[width=\columnwidth]{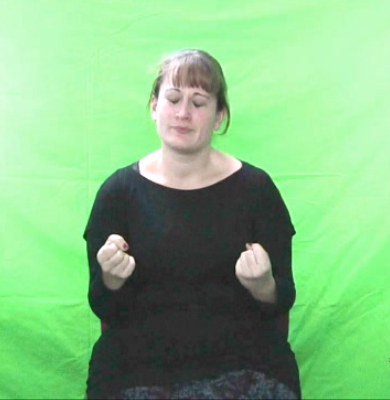}
                    \signlsf{DRIVE}
                \end{center}
            \end{minipage}
        }
        \vskip 2mm
        \centerline{
            \begin{minipage}{0.4\columnwidth}
                \begin{center}
                    \includegraphics[width=\columnwidth]{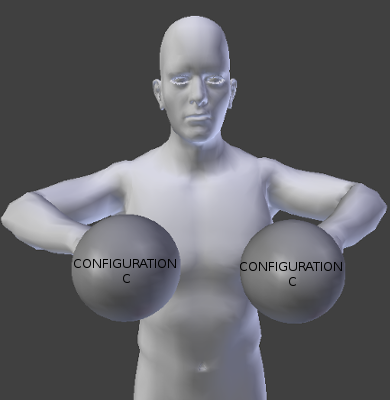}
                    $\mathtt{COMMON~STRUCTURE}$
                \end{center}
            \end{minipage}
        }
        \caption{Comparison of signs \signlsf{SCREEN} and \signlsf{DRIVE} sharing the same underlying structure}
         \label{methodology:signcomparison}
    \end{figure}

    Lambdas let us describe properties over sets of \ac{pdlsl} atoms instead of one. For example, Figure~\ref{methodology:signcomparison} shows two \ac{lsf} signs, \signlsf{SCREEN} and \signlsf{DRIVE}. Both can be described as instances of the same underlying common structure, characterized by both hands holding the same morphological configuration while being positioned opposite from one another.
        
    Their common base can be described by way of a lambda expression as shown in example~\ref{ex:comparison}.
    
    \begin{example}[\textbf{opposition} lambda expression]
        \begin{align*}
        hands\_config=&~\lambda c.(\mathcal{F}^\mathtt{~right}_{c}
                                   \wedge \mathcal{F}^\mathtt{~left}_{c})\\
        \mathbf{opposition}=&~\lambda c.(\mathtt{right}^\leftarrow_{\mathtt{left}} \wedge hands\_config(c))
        \end{align*}
        \label{ex:comparison}
    \end{example}
    \vspace{-6mm}

    In example \ref{ex:comparison}, $\mathcal{F}^\mathtt{~right}_{c}$ means that $\mathtt{right}$ holds configuration $c$. Atom $\mathcal{F}^\mathtt{~left}_{c}$ has the same meaning, but for the $\mathtt{left}$ hand. Atom $\mathtt{right}^\leftarrow_{\mathtt{left}}$ means that $\mathtt{right}$ hand lies in direction $\leftarrow$ with respect to $\mathtt{left}$, from the annotator's point of view. In this case we called our expression \textbf{opposition}, because both hands are in opposite horizontal positions from one another.

    Once we've defined the base structure, the \signlsf{SCREEN} and \signlsf{DRIVE} signs can be easily described in the database by passing the missing arguments to our lambda expression (as shown by example~\ref{ex:signs}).
    
    \begin{example}[\textbf{opposition}-derived signs]
        \begin{align*}
        \mbox{\signlsf{SCREEN}} =& ~\mathbf{opposition}(\mathtt{L\_FORM})\\
        \mbox{\signlsf{DRIVE}} =& ~\mathbf{opposition}(\mathtt{FIST\_FORM})
        \end{align*}
        \label{ex:signs}
    \end{example}
    \vspace{-6mm}
    
    In example~\ref{ex:signs}, $\mathtt{L\_FORM}$ is a morphological configuration of the hand where the thumb and the index fingers are held orthogonally. Similarly, $ \mathtt{FIST\_Form}$ is a configuration where hand is held as a closed fist. Here we just expressed that $\mathbf{opposition}$ will substitute each apparition of its first argument with either form, so as to define two distinct signs.
    
    We could also have described both signs as standalone, independent formulae. However, by describing the common structures across different signs, we are able to cope better with incomplete information in recognition. For example, a generic \textbf{opposition} structure with free variables will correctly hit in states where we can recognize hand positions but no hand configurations (as it's often the case). This immediately derives into a list of possible signs that could be later reduced with either further processing or with user interaction. In this scenario, standalone formulae for \signlsf{SCREEN} and \signlsf{DRIVE} wouldn't be found, since only using position information isn't enough to tell them apart.
    
\section{Detailed Framework Architecture}
    \label{sec:systemarch}

\begin{figure}[htb]
        \begin{tikzpicture}[auto, node distance=1.6cm, >=latex', inner sep=0pt]

            \tikzstyle{tool} = [fill=blue!20, rectangle, minimum width=1.4cm, minimum height=1.4cm]
            \tikzstyle{resource} = [fill=red!20, circle, minimum width=1.4cm]
            \tikzstyle{output} = [coordinate, node distance=1.6cm]
            \tikzstyle{tag} = [text width=4.6em, font=\scriptsize\sffamily, align=center]

            \node[resource] (corpus) {};
            \node[tool, right of=corpus](tracking){};
            \node[resource, below of=tracking, minimum width=1.4cm](segmentation){};
            \node[tool, right of=segmentation](modeliser){};
            \node[tool, right of=modeliser](checker){};
            \node[resource, above of=modeliser](model){};
            \node[resource, above of=checker](formules){};
            \node[resource, right of=checker](lexicaldata){};

            \node[tag, right=10pt of corpus, anchor=east](corpustag) {Corpus};
            \node[tag, right=10pt of tracking, anchor=east](trackingtag){Tracking and Segmentation Module};    
            \node[tag, right=10pt of segmentation, anchor=east](segmentationtag) {Key postures~\& transitions};
            \node[tag, right=10pt of modeliser, anchor=east](modelisertag){\pdlsl \\ Model Extraction Module};    
            \node[tag, right=10pt of checker, anchor=east](checkertag) {\pdlsl \\ Verification Module};
            \node[tag, right=10pt of model, anchor=east](modeltag){\pdlsl \\Graph};
            \node[tag, right=10pt of formules, anchor=east](formulestag){\pdlsl \\ Formulae DB};
            \node[tag, right=10pt of lexicaldata, anchor=east](lexicaldatatag){Verified Properties};
               
            \path[->] (corpus) edge (tracking)
                      (tracking) edge (segmentation)
                      (segmentation) edge (modeliser)
                         (modeliser) edge (model)
                      (model) edge (checker)
                      (formules) edge (checker)
                      (checker) edge (lexicaldata);
        \end{tikzpicture}
        \caption{Block diagram of a generic \ac{pdlsl}-based \ac{sl} lexical structure recognition system}
        \label{fig:block_diagram}
    \end{figure}
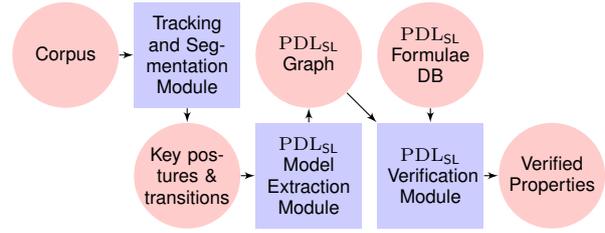

    The objective of the system is to take an untreated \ac{sl} video input, either in real time or not, and return a set of satisfied \ac{pdlsl} formulae. Moreover, the system has to return a \ac{pdlsl} model representing any relevant information contained in the video as a \ac{lts}. This can only be fulfilled by adapting the modeling process on-the-fly to the specific characteristics of our data. To achieve this end, our framework generalizes the original architecture proposed by~\cite{curiel_logic_2013}, shown in Figure~\ref{fig:block_diagram}, so as to enable module swapping depending on the technical needs presented by the inputs. 
    
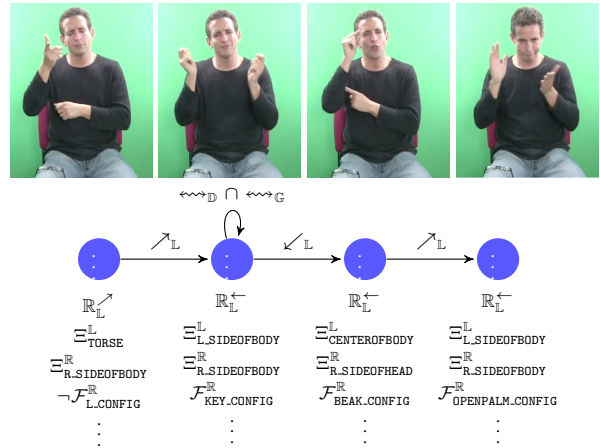
\begin{figure}[htb]
           \centering
           \begin{minipage}[b]{0.98\columnwidth}
               \centering
               \includegraphics[width=0.23\columnwidth]{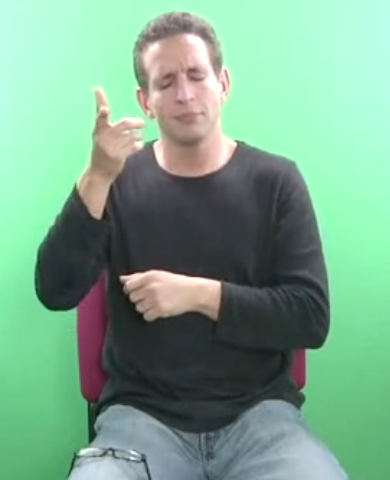}
               \includegraphics[width=0.23\columnwidth]{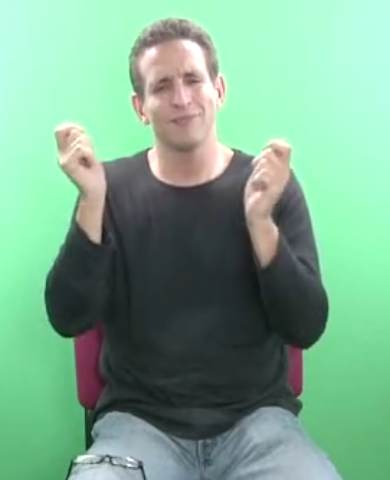}
               \includegraphics[width=0.23\columnwidth]{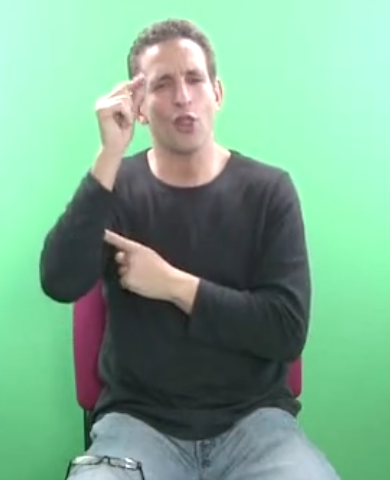}
               \includegraphics[width=0.23\columnwidth]{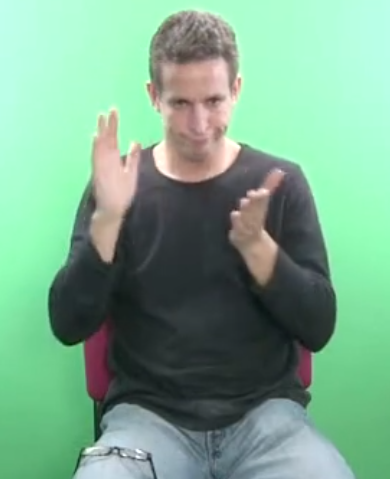}
           \end{minipage} \\
           \begin{minipage}[b]{.9\columnwidth}
           \centering
               \begin{tikzpicture}[->,>=stealth',shorten >=1pt,auto, font=\scriptsize]
                  \tikzstyle{every state}=[circle, fill=blue!65, draw=none, text=white,
                  text width=.2cm, inner sep=0pt, minimum size=10pt, node distance=1.75cm]
                  \tikzstyle{invisible}=[fill=none,draw=none,text=black, node distance=.4cm]
                  
                  \node[state] (s0) {$\vdots$};
                  \node[invisible] (s0tag1) [below=0.05cm of s0] {$\mathbb{R}^\nearrow_{\mathbb{L}}$};
                  \node[invisible] (s0tag2) [below of=s0tag1] {$\Xi^\mathbb{L}_{\mathtt{TORSE}}$};
                  \node[invisible] (s0tag3) [below of=s0tag2] {$\Xi^\mathbb{R}_{\mathtt{R\_SIDEOFBODY}}$};
                  \node[invisible] (s0tag4) [below of=s0tag3] {$\neg\mathcal{F}^{\mathbb{R}}_{\mathtt{L\_CONFIG}}$};
                  \node[invisible] (s0tag7) [below of=s0tag4] {$\vdots$};
                  \node[state] (s1) [right of=s0] {$\vdots$};
                  \node[invisible] (s1tag1) [below=0.05cm of s1] {$\mathbb{R}^\leftarrow_{\mathbb{L}}$};
                  \node[invisible] (s1tag2) [below of=s1tag1] {$\Xi^\mathbb{L}_{\mathtt{L\_SIDEOFBODY}}$};
                  \node[invisible] (s1tag3) [below of=s1tag2] {$\Xi^\mathbb{R}_{\mathtt{R\_SIDEOFBODY}}$};
                  \node[invisible] (s1tag4) [below of=s1tag3] {$\mathcal{F}^{\mathbb{R}}_{\mathtt{KEY\_CONFIG}}$};
                  \node[invisible] (s1tag7) [below of=s1tag4] {$\vdots$};
                  \path[->] (s0) edge node {$\nearrow_{\mathbb{L}}$} (s1);
                  \path[->] (s1) edge[loop above, distance=.5cm] node {$\leftrightsquigarrow_\mathbb{D} \cap \leftrightsquigarrow_\mathbb{G}$} (s1);
                  \node[state] (s2) [right of=s1] {$\vdots$};
                  \node[invisible] (s2tag1) [below=0.05cm of s2] {$\mathbb{R}^\leftarrow_{\mathbb{L}}$};
                  \node[invisible] (s2tag2) [below of=s2tag1] {$\Xi^\mathbb{L}_{\mathtt{CENTEROFBODY}}$};
                  \node[invisible] (s2tag3) [below of=s2tag2] {$\Xi^\mathbb{R}_{\mathtt{R\_SIDEOFHEAD}}$};
                  \node[invisible] (s2tag4) [below of=s2tag3] {$\mathcal{F}^{\mathbb{R}}_{\mathtt{BEAK\_CONFIG}}$};
                  \node[invisible] (s2tag7) [below of=s2tag4] {$\vdots$};
                  \path[->] (s1) edge node {$\swarrow_{\mathbb{L}}$} (s2);
                  \node[state] (s3) [right of=s2] {$\vdots$};
                  \node[invisible] (s3tag1) [below=0.05cm of s3] {$\mathbb{R}^\leftarrow_{\mathbb{L}}$};
                  \node[invisible] (s3tag2) [below of=s3tag1] {$\Xi^\mathbb{L}_{\mathtt{L\_SIDEOFBODY}}$};
                  \node[invisible] (s3tag3) [below of=s3tag2] {$\Xi^\mathbb{R}_{\mathtt{R\_SIDEOFBODY}}$};
                  \node[invisible] (s3tag4) [below of=s3tag3] {$\mathcal{F}^{\mathbb{R}}_{\mathtt{OPENPALM\_CONFIG}}$};
                  \node[invisible] (s3tag7) [below of=s3tag4] {$\vdots$};
                  \path[->] (s2) edge node {$\nearrow_{\mathbb{L}}$} (s3);
                \end{tikzpicture}
           \end{minipage}
           \caption{Example of modeling over four automatically identified frames as possible key postures}
           \label{fig:modeling}
       \end{figure}

    In the original version, a {\em Tracking and Segmentation} module uses the raw data of an automatic hand-tracker on 2D corpora, like the one presented by~\cite{gonzalez_robust_2011}, and returns a list of time-intervals classified either as {\em holds} or {\em movements}. The aforementioned interval list is passed to the {\em Model Extraction Module}, which translates each {\em hold} and {\em movement} into a time-ordered \ac{lts}. In the \ac{lts}, {\em holds} correspond to unique propositional states and {\em movements} map to transitions between states. An example of the resulting \ac{lts} is shown in Figure~\ref{fig:modeling}.
    
\begin{figure}[htb]
    \centerline{
        \includegraphics[width=.9\columnwidth]{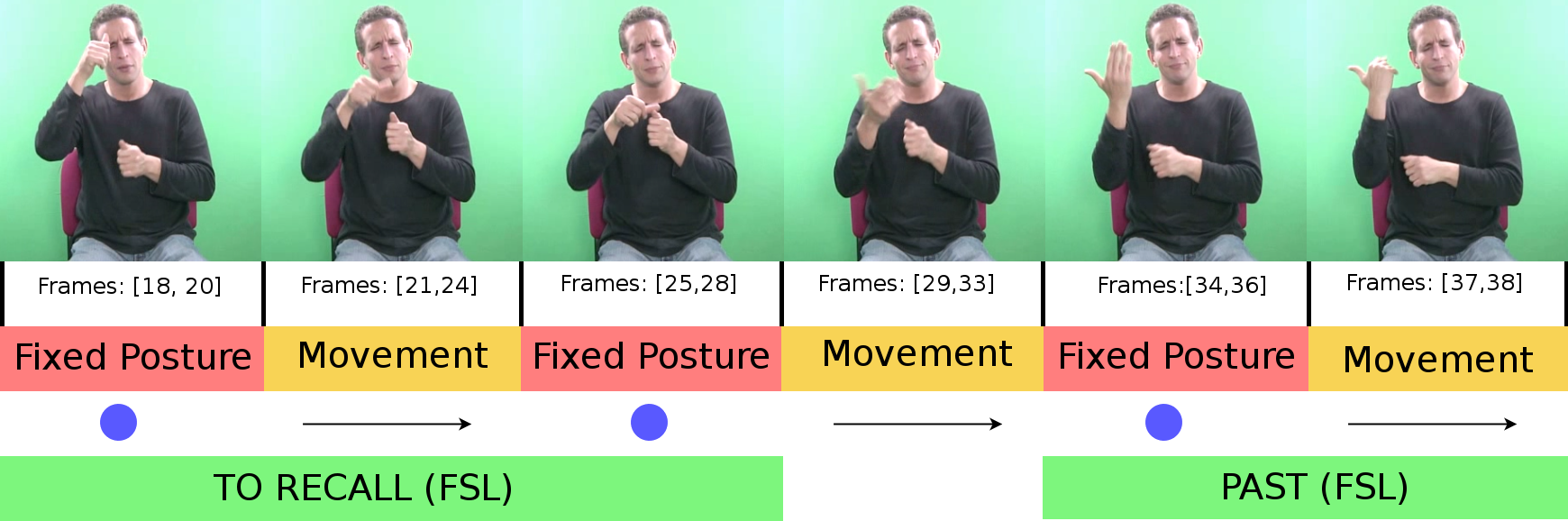}
        }
        \caption{Example of the different layers processed by  an automatic annotation system}
        \label{methodology:annotation}
    \end{figure}

    Finally, the {\em Verification Module} takes both the generated \ac{lts} and a database of \ac{pdlsl} formulae to determine which of them are satisfied in the model. As each formula corresponds to a formal description of a sign or property, the module can use logical satisfaction to verify if the property is present or not in the video. The complete process is shown in Figure~\ref{methodology:annotation}. Finally, the system maps each state where a formula is satisfied to its corresponding frame interval, so as to generate an annotation proposition.

     \begin{figure*}[b!]
         \includegraphics[width=2\columnwidth]{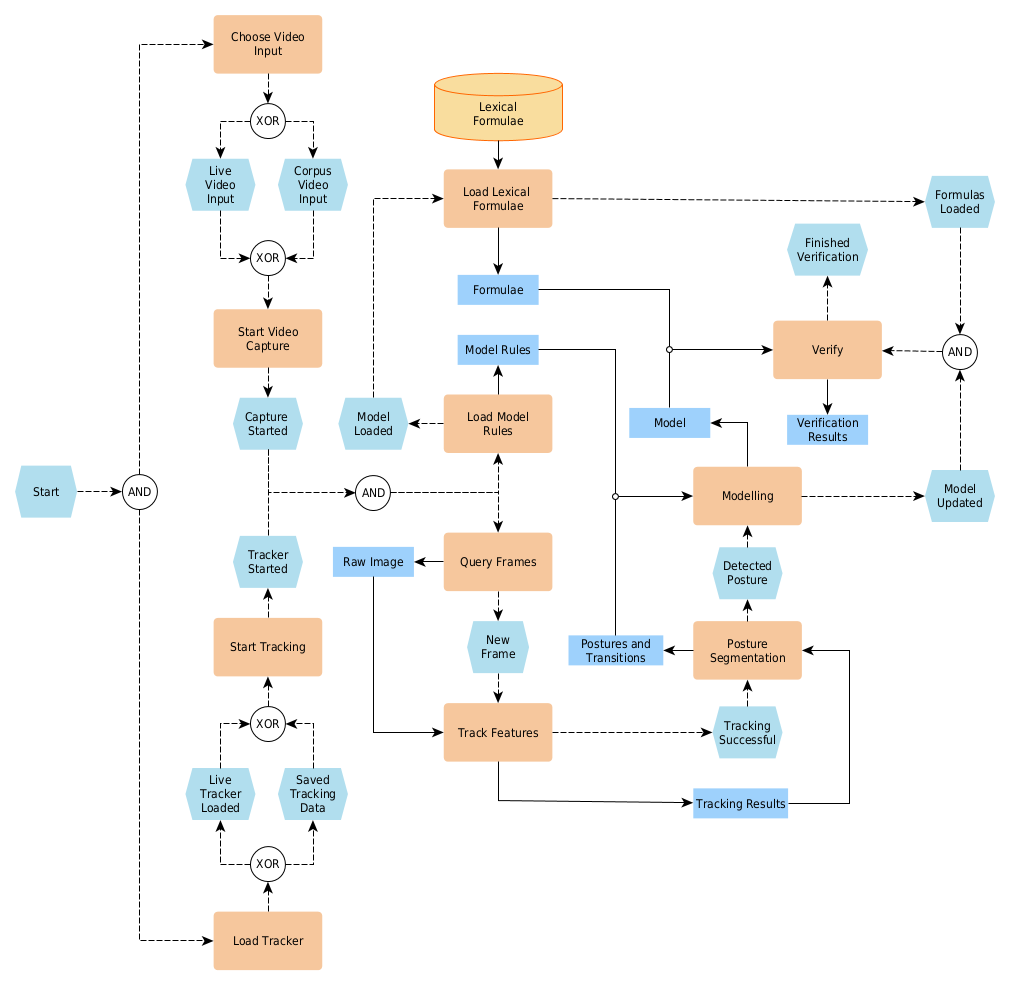}
         \caption{Information and control flow in the \ac{sl} annotation framework}
         \label{fig:architecture}
     \end{figure*}

\subsection{Observer Architecture Design}

    In order to be able to adapt dynamically to the particular needs of the input data, we devised the observer architecture shown in Figure~\ref{fig:architecture}.
    
    The main idea behind this design rests upon two axes:
    
    \begin{itemize}
    \item the possibility of using several tracking tools, adapted to different kinds of corpora;
    \item the generation of \ac{pdlsl} models consistent with the information generated by the different trackers.
    \end{itemize}
    
    Moreover, not only do models have to be consistent with every tracker but, as previously stated, not all trackers will give the same information nor track the same features. As such, the framework has to coordinate the loading of the proper modules depending on the corpus and the trackers. This process is entirely done by way of event-triggering. The same mechanism enables communication between modules by implementing \ac{mrsw} buffers, which allow every module to read their information but let only one of them write modifications. Each time a new modification is written in a \ac{mrsw} register, an event is issued system-wide to notify of the existence of new information. This event is then available to every module listening to that register's notifications. For the sake of compatibility, modules are obliged to implement an internal listening thread which can be subscribed to the communication channels of any other module.
    
    In general, the framework establishes development guidelines for the modules of the basic architecture, the one shown on Figure~\ref{fig:block_diagram}, so we can adapt them to specific cases without breaking compatibility. This is achieved by way of generic templates that implement the most basic functionalities of every module. These templates can later be extended to cover the specific cases arising in research; a developer can simply override the critical functionality in each template with their own code. Additionally, modules can register new events within the framework, so as to convey further information (if needed) for particular cases. As such, the system is capable of distributing self-contained, interchangeable, modules that can adapt to different situations.

    The execution process is also fairly straightforward. At the beginning a {\em Start} event is fired-up, prompting to load both a video stream and a tracker. This corresponds to the {\em Tracking and Segmentation Module} on the basic architecture (Figure~\ref{fig:block_diagram}). The system chooses between the compatible video inputs and pairs the selection with the proper tracker. This is done by reading the events sent out by the loading functions. Likewise, the model construction rules are loaded after a compatible set of video/tracking inputs has been selected. In this way, we can assure that the modeling algorithm will only take in account pertinent rules, those relying on the specific features we are tracking. This mechanism avoids generating models based on hand positions, for example, if our tracker is only capable of detecting non-manuals. Once a compatible set of modules is activated, the process can continue as proposed by~\cite{curiel_logic_2013}.
    
\section{Experimental Results}
    \label{sec:results}

    We obtained some preliminary results on the proposed framework by implementing the system's core and a set of minimal templates for each of the modules on Figure~\ref{fig:block_diagram}. The core contains the necessary data structures to represent both \ac{pdlsl} models and formulae, alongside the semantic rules necessary to acknowledge logical satisfaction.
        
    For the creation of the module templates, we considered two possible scenarios:
    
    \begin{itemize}
    \item the system is being used to annotate previously captured video corpora;
    \item a camera as going to be used as input for real-time sign recognition.
    \end{itemize}
    
    Furthermore, we had to consider two distinct cases when treating video; whether we had 2D or 3D information available for determining relationships between hands and body. For simplicity, we worked only with hand-tracking data. Nevertheless, the addition of non-manual trackers is also a possibility, since introducing new modeling rules for non-manuals follow the same principles of the 2D to 3D transition.
    
    Once all the framework tools were in place, we created a specific implementation for the 2D case, when tracking features over existing corpora.

    \subsection{Automatic Annotation in 2D Corpora}

    To obtain some initial results over real-world data, we developed the first modules based on the atoms originally presented with the \ac{pdlsl} language. Additionally, we created a property database made of \ac{pdlsl} formulae, adapted to be used with our tracking device. The database position in the architecture is shown in Figure~\ref{fig:architecture}, as the node {\em Lexical Formulae}. The formulae were exclusively constructed for the 2D case; this means that, for any other kind of tracking information, we would need to define new \ac{pdlsl} database with different properties. For tracking, we used the tracker developed by \cite{gonzalez_robust_2011}, which is capable of finding 2D positions of the hands and head over \ac{sl} video corpora. As for the \ac{sl} resources, we used an instance of the {\em DictaSign} corpus \cite{dictasign} as video input for our system.
    
    Since the used tracking tool is not adapted for real-time processing, the implemented tracking module just recuperates the previously calculated information from an output file. This is done sequentially, after each successful querying of a new video frame, to simulate real-time.
    
    To calculate the posture segmentation we used the method proposed by \cite{gonzalez_sign_2012}, which is based on measuring speed-changes. 
    
    Our \pdlsl ~description database contains four structures:
    
    \begin{description}
    \item[opposition.] $\lambda c.(\mathtt{right}^\leftarrow_{\mathtt{left}} \wedge hands\_config(c))$. Hands are opposite to each other, with the same configuration.
    \item[tap.] $\lambda s, w.(\neg\mathcal{T}^{s}_{w} \rightarrow [moves(s)\cup moves(w)] \mathcal{T}^{s}_{w} \rightarrow [\mathbf{skip};\mathbf{skip}] \neg\mathcal{T}^{s}_{w})$. Hand touches briefly the other hand, only for a single state.
    \item[buoy.] $\lambda s, posture.(posture~\wedge~[moves(s)^*] posture)$. The state of one hand remains the same over several states, regardless of the movements of the other hand.
    \item[head anchor.] $\lambda s, w, posture.(\mathbf{buoy}(s, posture)~\wedge~\mathcal{T}^\mathtt{head}_{w})$. One of the hands remains within the head region while the other signs.
    \end{description}
    
    The $posture$ variable denotes the propositional state of an articulator. The $moves(s)$ function can be interpreted as any action executed by articulator $s$. We omit the complete, formal definition of this operator for the sake of readability. 
    
    To measure the {\em hit} ratio of the system, we manually annotated the apparition of the described properties in one video within the corpora. Table~\ref{table:manual_annot} shows the quantity of observed apparitions of each property over the chosen video.

    \begin{table}[htb]
        \centerline{
        \begin{tabular}{|c|c|c|c|c|}
        \hline
        $\mathbf{\varphi}$ & \textbf{oppos.}& \textbf{buoy}&\textbf{tap}&\textbf{h. anch.} \\ \hline
         \textbf{Total} & 76 & 40 & 33 & 74 \\ \hline
        \end{tabular}}
        \caption{Manually annotated apparitions of property formulae on one video}
        \label{table:manual_annot}
    \end{table}
    
    For each signer, the system creates a model based only on the atoms specified by the modeling rules. It then uses the created model to verify every formula on-the-fly. The execution of our algorithm over the same video rendered the results shown in Table~\ref{table:predicted_annot}.
    
    \begin{table}[htb]
        \centerline{
        \begin{tabular}{|c|c|c|c|c|}
        \hline
        $\mathbf{\varphi}$ & \textbf{oppos.}& \textbf{buoy}&\textbf{tap}&\textbf{h. anch.} \\ \hline
        \textbf{Total} & 164 & 248 & 79 & 138 \\ \hline
        \end{tabular}}
        \caption{Total reported hits of property formulae on one video}
        \label{table:predicted_annot}
    \end{table}

    On Table~\ref{table:predicted_annot} we can see the total number of times each of the formulae were verified on the video, as returned by the system. We compare the human annotations with these results on Figure~\ref{results:results}.

    \begin{figure}[htb]
    
        \centerline{
            \begin{tabular}{|c|c|c|c|c|}
            \multicolumn{5}{c}{\textsc{Total Observations}}\\ \hhline{=====}
            $\mathbf{\varphi}$& \textbf{oppos.}& \textbf{buoy}&\textbf{tap}&\textbf{h. anch.} \\ \hline
            \textbf{By hand} & 76 & 43 & 33 & 74 \\ \hline
            \textbf{Automatic} & 164 & 245 & 79 & 138 \\ \hline
            \end{tabular}
        }
            
        \vskip 2ex 
        
        \begin{tabular}{c|c|c|c|c|c|}
         \multicolumn{2}{c}{}&\multicolumn{4}{c}{\texttt{Automatic}} \\ \hhline{~~|----|}
         \multicolumn{1}{c}{}&$\varphi$&\textbf{oppos.}&\textbf{buoy}&\textbf{tap}&\textbf{h. anch.}\\ \hhline{~-----}
         \multirow{5}{*}{\rotatebox{90}{\texttt{Human}}}
         &\textbf{oppos.}& \cellcolor{gray!30}67& 64& 10& 33\\ \hhline{~-----}
         &\textbf{buoy}& 22& \cellcolor{gray!30}40& 7& 17\\ \hhline{~-----}
         &\textbf{tap}& 15& 24& \cellcolor{gray!30}25& 11\\ \hhline{~-----}
         &\textbf{h. anch.}& 23& 50& 13& \cellcolor{gray!30}44\\ \hhline{~-----}
         & {\em False P.}& 37& 67& 24& 33\\ \hhline{~-----}
        \end{tabular}
        \caption{Formulae verification results}
        \label{results:results}
    \end{figure}

    Figure~\ref{results:results} shows data from both Tables~\ref{table:manual_annot} and ~\ref{table:predicted_annot}, 
    alongside a {\em matching table} where, for each property formula, we count the quantity of times it was verified on previously human-annotated frames. Each row represents the total number of human observed apparitions of one property, while each column represents the quantity of positive verifications reported by the system for each formula. For example, cell (\textbf{opposition}, \textbf{opposition}) shows the total number of times the \textbf{opposition} formula was correctly verified on human-annotated \textbf{opposition} frames. The next cell, (\textbf{opposition}, \textbf{buoy}), holds the number of times the \textbf{buoy} property was verified on human-annotated \textbf{opposition} frames. Positive verifications can overlap, {\em i.e.} the system could have verified two or more formulae over the same states of the model. Therefore, a single annotation could belong to different classes. The cells on the last row of the table correspond to {\em false positives}, reported detections that don't overlap with any human observation.

    Further analysis on the matching table is represented on Table~\ref{table:summary_results}, which shows the total number of correctly and incorrectly classified formulae, as well as the total mismatches.
    
    \begin{table}[htb]
    \begin{tabular}{cccc}
    \hline
    \multirow{2}{*}{$\varphi$}
    &\multicolumn{2}{c}{\textsc{Human Obs.}}&
    \multirow{2}{*}{\textsc{Erroneus Match}}\\ \hhline{~--~}
    &\textsc{Hit}&\textsc{Miss}\\ \hhline{====}
    \textbf{opposition}& 67&  9& 107\\
    \textbf{buoy}      & 40&  3&  46\\
    \textbf{tap}       & 25&  8&  50\\ 
    \textbf{h. anchor} & 44& 30&  86\\
    \end{tabular}
    \caption{Per-formula summary of the total number of observations found, missed and erroneously classified observations}
    \label{table:summary_results}
    \end{table}
    
    The results show a high recognition rate for \textbf{opposition}, \textbf{buoy} and \textbf{tap}, but also a high quantity of misclassification hits and false positives. Most of the erroneous hits are due to the definitions of the properties themselves. Take, for example, \textbf{opposition} and \textbf{buoy} properties. In the video, some of the states satisfying a \textbf{buoy} could easily be classified as \textbf{opposition}. When this happens, the only thing that differentiates them, if we only have tracking information, is their movement: if a hand is moving is a \textbf{buoy}, otherwise is an \textbf{opposition}. Even though this is not always the case, sometimes the situation arises and the system confuses these properties for one another; if some of the movements of the hands are too fast, or not ample enough, when performing a \textbf{buoy}, the system interprets them as a static posture, therefore classifying some of the internal states of the \textbf{buoy} as \textbf{opposition}. This, however, doesn't impede finding the buoy, since the definition of \textbf{buoy} specifies, from the beginning, an arbitrary number of internal states, hence not affected by having found one instead of two distinct states. The opposite case might also arise, when a short involuntary movement, is interpreted by the system as an intended action instead of noise, hereafter classifying an \textbf{opposition} as a \textbf{buoy}, or even as two sequential \textbf{oppositions}. Similar arguments can be made for \textbf{tap} and \textbf{head anchor}, where movement thresholds alone can affect the form and the quantity of states on the \ac{lts}. In the future, we expect that adding new information will reduce the quantity of misclassified data, specially because this will result in a more fine-grained model from the beginning.
    
    At this stage, the system returns a list of proposed properties as result of the verification phase. What the numbers on Table~\ref{table:summary_results} mean is that, in most cases, the proposed annotation will almost never return single properties but rather sets of properties. This may not be a problem with simple formulae like the ones described, but would be problematic with complete sign descriptions; there is such thing as too much information. In that case, we would need a human being to complete the classification process. This points out the need or a higher level module in charge of cleaning the annotation proposal by way of machine learning techniques.
    
    Finally, most of the false positives that don't correspond to any overlap with human observations were caused by signer's movements without communication intent. For example, some \textbf{opposition} properties were found when a signer crossed his arms, when his hands were posed over his knees or when he assumed other natural repose positions. Similarly, some co-articulatory movements created chains of states that satisfied the formulae for \textbf{buoy} or \textbf{tap}. These cases could also be reduced with help of a higher level module or a human expert.
    
    \subsection{Extending to 3D}
    
\begin{figure}[htb]
    \centerline{
    \includegraphics[width=.9\columnwidth]{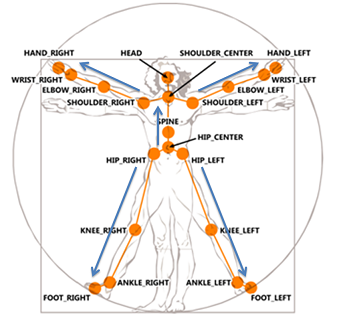}
    }
    \caption{Points tracked by the Kinect device~\cite{msdn_2013}}
    \label{fig:kinectjoints}
    \end{figure}

    Currently, we are extending the system to model features tracked in 3D. We have already extended the framework to process data returned by the Kinect~\cite{msdn_2013}, a motion sensing device capable of tracking 3D positions on several body articulations. Figure ~\ref{fig:kinectjoints} shows the points that can be tracked by using the Kinect with it's official development kit.
    
    For the moment, we have been able to reuse the same modeling rules that we implemented for the 2D case; mainly, we have used the Kinect tracker to obtain 3D position data of hands and head, and we have projected this information in 2D. This lets us create the same kind of models we build from corpora. However, the variety of the tracked articulations and the 3D capabilities of the sensor, call for the definition of more complex atoms and lambda properties, as well as 3D descriptions of individual signs. As of 2014, work is still ongoing on the matter and has not been properly evaluated. Nevertheless, we considered important to point out that we can already exchange trackers if needed, so as to showcase the flexibility of our framework.

\section{Conclusions}
    \label{sec:conclusions}
    
    Here we have presented an automatic annotation framework for \ac{sl} based on a formal logic. The system lets us represent \ac{sl} video inputs as time-ordered \acp{lts} by way of \ac{pdlsl}, a multi-modal logic. We have shown that it is possible to use the resulting graph to verify the existence of common lexical structures, described as logical formulae. Furthermore, the framework gives us the necessary tools to adapt the model generation for different corpora and tracking technologies.

    From the point of view of recognition, we noticed that the quality of the tracking tools is of utmost importance for both formula definition and model generation. The low presence of information and high levels of noise immediately took a toll on verification; in some cases, we lacked enough information to distinguish between intended movements and noise. In turn, this resulted on high rejection rates of what would otherwise be considered {\em hit} frames.
    
    Similarly, we noticed that modeling can be affected by the presence of low information, which can render states indistinguishable. For instance, without hand configurations every state satisfying \textbf{opposition} is, effectively, the same state. Therefore, every formula sharing the same \textbf{opposition} base would be satisfied on that single state. This could gravely affect the system's performance; in the worst case, all states could satisfy all formulae. On the other hand, a too fine-grained model can lead to a \ac{lts} that replicates the same problems we have in synthesis-oriented descriptions. In that case, we would need very specific formulae (with near to perfect \ac{sl} corpora) to achieve any identification at all. Similarly, formula creation can't be neither too broad nor too specific, if we want to minimize the quantity of imperfect matches. Anyhow, one of the advantages we have by using a logical language is that we can control the granularity of information simply by defining or discarding atoms, which opens the door to the use of algorithmic techniques to control information quantity. 
    
    From the point of view of the implementation, the results of the 2D experiments show that further effort has to be put on integrating new sources of information to the system, especially if we want avoid false positives. Even though the system is in place and works as expected, the high quantity of erroneous hits reflects the gravity of the problems we can have with indistinguishable states. Further comparisons have to be done once the system completely incorporates 3D modeling, so as to measure the effective impact of additional information on verification.
    
    Future work in recognition will be centered on implementing machine learning techniques to improve verification. Using data analysis to find relationships between detected structures, could lead us to better results even in suboptimal environments. Additionally, we would like to integrate communication with user level software like the one presented by~\cite{dubot_improvements_2012}, a manual annotation tool. This could lead to other possible uses of the framework as engine for higher applications, such as dictionary searching or even for automatic creation of sign description databases from \ac{sl} videos.
    
    Further analysis will also target the building blocks of the language, by changing the semantic definitions of \ac{pdlsl} operators to better suit \ac{sl}. Changes to its syntax are also to be expected, in an effort to ease the development of extensions for different trackers and simplify descriptions. Finally, we want to steer further into 3D representation and the inclusion of non-manual features, important stepping stones towards higher level language processing.
    
\bibliographystyle{lrec2014/lrec2014}
\bibliography{slwshop14}

\end{document}